\documentclass[conference]{IEEEtran}
\IEEEoverridecommandlockouts

\usepackage{fancyhdr}

\usepackage{cite}
\usepackage{amsmath,amssymb,amsfonts}
\usepackage{algorithmic}
\usepackage{graphicx}
\usepackage{textcomp}
\usepackage{eso-pic}
\usepackage{xcolor}

\usepackage[ruled,vlined]{algorithm2e}
\usepackage{array}

\usepackage{multirow}

\usepackage{flushend}

\ifCLASSOPTIONcompsoc
    \usepackage[caption=false, font=normalsize, labelfont=sf, textfont=sf]{subfig}
\else
\usepackage[caption=false, font=footnotesize]{subfig}
\fi

\def\BibTeX{{\rm B\kern-.05em{\sc i\kern-.025em b}\kern-.08em
    T\kern-.1667em\lower.7ex\hbox{E}\kern-.125emX}}

\makeatletter

\usepackage{eso-pic}
\ifCLASSINFOpdf
\else
\fi
\begin{document}

\title{Identifying Alzheimer Disease Dementia Levels Using Machine Learning Methods}

\makeatletter
\newcommand{\linebreakand}{%
  \end{@IEEEauthorhalign}
  \hfill\mbox{}\par
  \mbox{}\hfill\begin{@IEEEauthorhalign}
}
\makeatother


\author{
\IEEEauthorblockN{Md Gulzar Hussain}
\IEEEauthorblockA{School of Computer and Artificial Intelligence,\\
Changzhou University\\
Changzhou, Jiangsu, China\\
Email: gulzar.ace@outlook.com}   

\and

\IEEEauthorblockN{Ye Shiren}
\IEEEauthorblockA{School of Computer and Artificial Intelligence,\\
Changzhou University\\
Changzhou, Jiangsu, China\\
Email: yes@cczu.edu.cn}
}

\maketitle



\begin{abstract}
\boldmath
Dementia, a prevalent neurodegenerative condition, is a major manifestation of Alzheimer's disease (AD). As the condition progresses from mild to severe, it significantly impairs the individual's ability to perform daily tasks independently, necessitating the need for timely and accurate AD classification. Machine learning or deep learning models have emerged as effective tools for this purpose. In this study, we suggested an approach for classifying the four stages of dementia using RF, SVM, and CNN algorithms, augmented with watershed segmentation for feature extraction from MRI images. Our results reveal that SVM with watershed features achieves an impressive accuracy of 96.25\%, surpassing other classification methods. The ADNI dataset is utilized to evaluate the effectiveness of our method, and we observed that the inclusion of watershed segmentation contributes to the enhanced performance of the models.

\end{abstract}



\section{Introduction}
The brain, being one of the highly vital and complicated organs in a human body, is responsible for a wide range of essential functions such as intellectual invention, problem-solving, thought, judgment, creativity, and memories. However, when an individual develops dementia, their cognitive abilities become impaired. The majority of all dementia sufferers have Alzheimer's disease (AD), making it the most prevalent reason of dementia. AD gradually destroys brain cells, resulting in disconnection from surroundings, loss of recognition of loved ones, inability to recall childhood memories, familiar faces, and even basic procedures. Alarmingly, the number of individuals affected by AD is projected to increase significantly in the coming decades. It is estimated that by 2050, there will be over 150 million AD sufferers worldwide, up from 50 million in 2020. This rapid increase in AD cases poses a significant burden on patients, families, and healthcare systems, unless there are advancements in prevention measures utilizing modern medical technologies. The need for research and innovation in AD prevention and treatment is crucial to address this growing global health challenge. Efforts to develop effective medications, leveraging contemporary medical technologies, are essential to combat the rising prevalence of AD and alleviate the impact it has on individuals, families, and healthcare systems worldwide.


According to the estimates, there are 32, 69, and 315 million people worldwide who have prodromal AD, AD dementia, and preclinical AD, respectively. Combined, they made up 416 million people on the AD spectrum or 22\% of all those 50 and older \cite{gustavsson2023global}. The burden of dementia grew globally between 1990 and 2019, increasing by 147.95\% and 160.84\%, correspondingly, in frequency and prevalence. In order to deal with the rising incidence of dementia, we should pay focus to the elderly society, prioritize programs that focus on dementia health conditions, and create plans of action \cite{li2022global}. Considering millions of people struggling with dementia globally, the prevalence of dementia has a significant necessity. The indications of AD and those of vascular dementia (VD) or ordinary aging intersect, making the identification of AD problematic \cite{castellazzi2020machine}. Premature and precise identification of Alzheimer's disease (AD) is crucial for patient care, prevention, and treatment. Several research initiatives aim to utilize imaging of brain techniques, including the magnetic resonance imaging (MRI), for identifying AD. Neuroimaging methods like structural MRI (sMRI), amyloid PET, functional MRI (fMRI), fluorodeoxyglucose positron emission tomography (FDG-PET) imaging, and diffusion tensor imaging (DTI) have shown promise in evaluating abnormal brain changes associated with AD. By leveraging these advanced neuroimaging methods, researchers and clinicians can monitor the progression of AD, enable earlier and more accurate diagnosis, and facilitate appropriate patient care, therapy, and prevention strategies. \cite{odusami2021analysis}. fMRI has aided AD investigators in assessing functionally active areas when performing a task to identify AD early in comparison to other neuro-imaging methods. 
The hippocampus, a crucial brain region responsible for creating and retaining new memories, is one of the earliest areas to be influenced by Alzheimer's disease. As Alzheimer's progresses, the hippocampus undergoes structural changes, resulting in a decline in the brain's ability to create and retrieve memories. Advanced imaging techniques, such as 3D imaging through Alzheimer's MRI, can provide visual representations of the hippocampus, depicting the number of cells present and its size. By utilizing 3D imaging, researchers and clinicians can gain a clearer understanding of the extent of structural changes in the hippocampus caused by Alzheimer's \cite{RoleofMRIs:online}. Mild cognitive impairment (MCI), which is often considered as a intermediate stage in the middle of normal cognitive aging and Alzheimer's disease, can be detected using MRI imaging. 

Modern developments in image processing have helped to improve quickly in developing modern methods. The use of image analysis has grown, particularly in the realm of healthcare. In comparison to the current approaches, deep learning offers more in the way of time and effectiveness. Great advancements have been achieved in the employment of machine learning (ML) and deep learning (DL) methodologies to the determination of AD, especially utilizing magnetic resonance imaging (MRI), which has been facilitated by the availability of open AD-related databases. In contrast to conventional ML techniques, DL has emerged as a paradigm-shifting tool in recent years. Since DL can train immediately from images using neural networks, this has simplified the procedure without involving specialists in extracting features. This eliminates the need for feature extraction to be carried out physically and separately from the classifiers \cite{ji2019early}. While there are many other ways to extract features, deep learning (DL) using multilayered ANNs, also known as deep neural networks (DNNs), provides a potential substitute by making feature extrication automated. Earlier feature extraction has been heavily used in earlier studies to lower the dimensions of actual-world inputs like MRI and expressed gene data \cite{tanveer2020machine}. Although this method can provide models with great accuracy, it can also result in models that do not take the fresh correlations found in the raw data into account. As a step of the learning process of the model, innovative methods to the structure of DNNs, including convolutional neural networks (CNNs) and, spiral neural networks (SNNs) enable the dimensions of the training dataset to be decreased, enabling the system to engage cooperatively with the raw input \cite{wen2020convolutional}. Convolutional neural networks (CNNs) has recently attained extremely high quality and reliability in image classification \cite{ji2019early}. 

The watershed algorithm using the conception of topography, which is fundamentally a region-growing technique, differs in that it starts development from the local minimum throughout the images. There will be a bunch of pseudo-minimum because of the impact of the darker disturbance and darker roughness in the photo, and these pseudo-minimums result in the associated pseudo-water basins in the picture. Hence, the watershed method separates out every pseudo-real minimum's minimum amount into a distinct region \cite{zhang2011marker}. The target of this study is to evaluate watershed algorithm-based  extraction of features from MRI for the automatic detection and categorization of four dementia levels of Alzheimer's disease depending on the superior execution of support vector machine (SVM), random forest (RF), and convolutional neural network (CNN) approach in a variety of images classification challenges. Alzheimer's disease utilizing MRI images is being diagnosed using RF, SVM, and CNN classifier models, and the effectiveness of the models was evaluated between them. The investigation questions listed below are the focus of the study's objectives. 

\begin{enumerate}
  \item Is it effective to classify Alzheimer's diseases using MRI brain images to employ the watershed algorithm feature extraction approach with RF, SVM and CNN classifiers? 
  \item Which classifier will provide us with higher classification results in this study? 
  \item How this method performs with low resolution images in this study?
\end{enumerate}

The remaining of this article is arranged as follows: Section \ref{Rel} covers prior studies of AD diagnosis and classification. The methods for building and evaluating the suggested models is presented in Section \ref{Medhod}. The experimental and evaluation results are described in Section \ref{Res} and section \ref{con} closes the paper and discusses further research.

\section{Related Work}\label{Rel}
Depending on high-dimensional data obtained from various types of neuroimaging biomarkers, such as MRI, numerous machine learning techniques have been exhibited to assist in the diagnosis of Alzheimer's. Throughout this section, recent works that employed traditional DL and ML methods in AD diagnosis and prediction methods are addressed. 

The authors of the work \cite{alsaeed2022brain} suggested employing a previously trained CNN deep learning system called ResNet-50 as an automated features extraction technique for MRI image-based Alzheimer's disease diagnosis. The effectiveness of a CNN containing traditional SVM, Softmax, and RF was then assessed using several metrics, such as accuracy, and it was found to perform better, with a range of accuracy for classifiers using the MRI ADNI data of 85.7\% to 99\%. The correctness of CNN is 97.6\%, that is 10.96\% greater than the Spinl-Net technique, when using the XAI approach utilized in the study \cite{kamal2021alzheimer}, local interpretative model agnostic explanation (LIME), for a straightforward human explanation. Support vector classifier offers greater accuracy when assessing gene expression data than other methods. The most crucial genes for a certain AD patient were identified from the gene expression data, and LIME demonstrates how genes were chosen for that patient. The adoption of the weighted map for both the 2 young radiologists in the matter of EOAD vs. FTD dramatically enhanced diagnostic performance in article \cite{chague2021radiological}. The increase in diagnosis accuracy was over 10 percent. The study \cite{bharanidharan2020improved}'s goal was to use an enhanced chicken swarm optimization (ICSO) approach to classify MRI pictures as demented (DM) and non-demented (ND). The traditional chickens swarm optimization method has an efficiency of 52.13\% using statistical features and 52.99\% without, meanwhile, the ICSO without statistical features has accurate results of 86.32\%. The approach used in paper \cite{haque2021deep} can evaluate many classes in a singular installation with fewer tagged training specimens and little previous domain knowledge than earlier methods by resolving the bottleneck problem. Authors of \cite{hazarika2022experimental} suggested a depth-wise convolution operation in place of the existing DenseNet-121 architecture's convolution operation to reduce completion time. The model's efficiency was also enhanced by the new framework, on average by 90.22\%.

\begin{figure*}[htb]
    \centering
  \includegraphics[width=18cm,height=9.5cm]{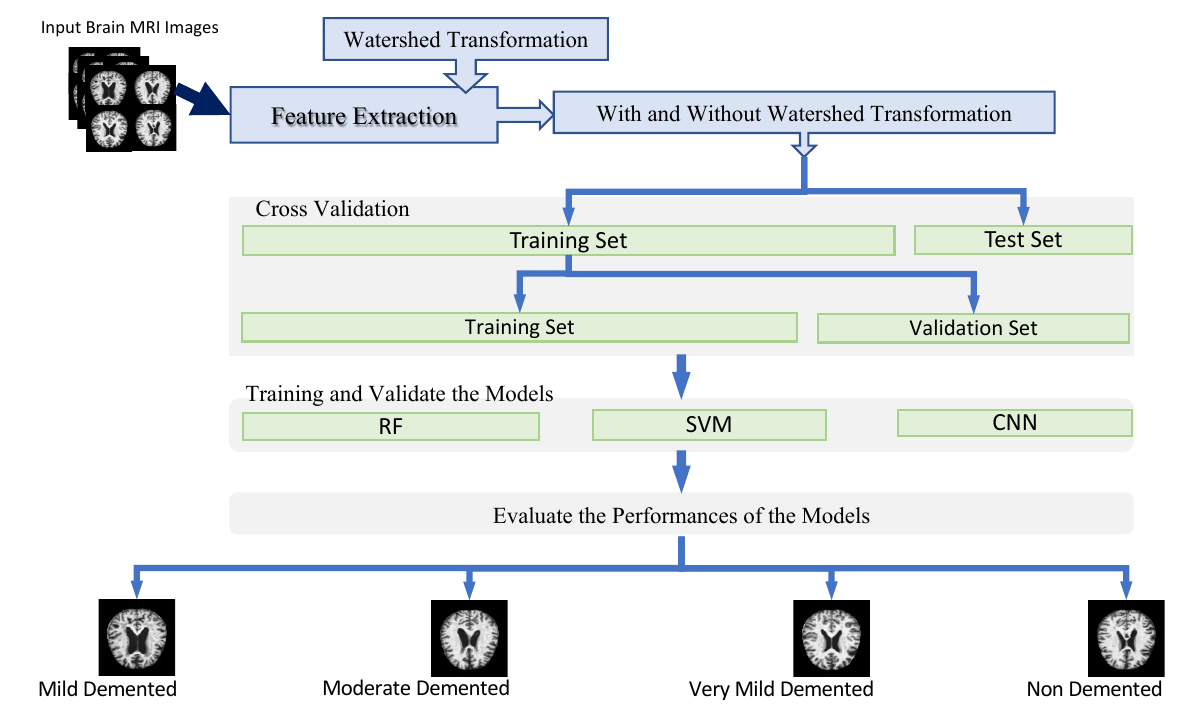}
    \caption{Flow Diagram of the Suggested Method in this Research Work}
    \label{systemFlow}
\end{figure*}

The proposed methodology in article \cite{abdulazeem2021cnn} correctly categorized instances of constitutionally normal (CN) and Alzheimer's disease (AD) employing the Alzheimer's disease Neuroimaging Initiative (ADNI) dataset having accuracy scores of 99.6 percent, 99.8 percent, and 97.8 percent, correspondingly. In multi-classification investigations, the proposed architecture achieved a categorization accuracy of 97.5 percent on the ADNI dataset. In order to categorize the several levels of Alzheimer's Disease using the magnetic resonance imaging (MRI) pictures, VGG16 influenced CNN (VCNN) and Deep Convolution Neural Network (DCNN) architectures had been established in article \cite{suganthe2020diagnosis}. The findings of experimentation on an ADNI dataset demonstrate that the proposed approaches achieved outstanding accuracy. To recognize the phases of dementia using MRI, a DEMentia NETwork (DEMNET) is proposed in article \cite{murugan2021demnet}. Using the Kaggle dataset, this DEMNET acquires an accuracy score of 95.23 percent, an area under curve (AUC) of 97\%, and a Cohen's Kappa score value of 0.93. A deep learning-based algorithm is suggested in the research \cite{odusami2021analysis} which can forecast MCI, late MCI (LMCI), early MCI (EMCI), and AD. In their study for testing, the 138-subject ADNI fMRI dataset was employed. For the AD vs. EMCI, AD vs. LMCI, and EMCI vs. MCI classifying settings, the fine tuned ResNet-18 network acquired an accuracy score of classification of 99.99\%, 99.95\%, and 99.95\%, correspondingly. In the work \cite{miltiadous2021alzheimer}, the categorization of undertaken EEG signals from FTD and AD patients was examined across six supervised machine-learning approaches. The accuracy ratings for the suggested approach were 86.3\% for detection of FTD using random forests and 78.5\% for detection of AD using decision tree. The determination proof from a three year clinical follow up was compared with ML predictions in the paper \cite{castellazzi2020machine}. The most effective method for differentiating AD from VD was ANFIS, which used a minimal feature pattern to obtain a accuracy of classification greater than 84\%. Using the most frequent T1-weighted sequencing magnetization produced fast accumulation with gradient echo (MPRAGE), the presented structure in \cite{hu2021deep} produced an accuracy of 91.83\%. In order to extract certain useful features from PSD images and carry out the associated two and three-dimensional classification problems, a modified CNN including one processed element of convolution, pooling layer, and Rectified Linear Units (ReLu) is constructed in \cite{ieracitano2019convolutional}. It's accuracy obtained in binary classification is 89.8\%, while in three classifications, it is 83.3\%. In the study \cite{chandaran2022deep}, research on disease detection over the past five years utilizing transfer learning based on deep learning for only brain MRI scans is conducted. This research suggests that a composite approach focused on transfer learning provided over an accuracy of 90\% in the majority of scenarios with a little training period. 

According to the literature, there are many methods for classifying AD employing machine and deep learning. Nevertheless, the multi-label AD classification's significant model parameters and class imbalance remain a problem. Feature extraction by the watershed algorithm is employed and for classification, RF, SVM and CNN model with fewer parameters is developed \& compared to address this issue and forecast four levels of dementia that can guide to AD in this research work.

\section{Proposed Methodology}\label{Medhod}
The primary objective of this research is to evaluate and improve the performance of the classifier of MRI images for the classification of dementia levels of Alzheimer's disease using RF, SVM, and CNN. Data gathering, pre-processing, Deep Learning based performance tuning and classifying, and impact assessment are all part of the research technique. The MRI data were obtained from a famous Alzheimer's disease database \cite{KaggleDatasetADNI}. Figure \ref{systemFlow} depicts the suggested model's process flow chart.

\subsection{Data Collection}
For this study, we used MRI scans of the brain acquired from the ADNI (Alzheimer’s Disease Neuroimaging Initiative) dataset \cite{jack2008alzheimer} and downloaded from \cite{KaggleDatasetADNI}. The dataset is comprised of 6400 pre-processed MRI (Magnetic Resonance Imaging) images of four types. All of the pictures have been scaled to 128 x 128 pixels. Sample images of each categories are given in figure \ref{sampleImages}.

\begin{figure}[htb]
    \centering
  \subfloat[Mild Demented.\label{md}]{%
       \includegraphics[width=0.23\linewidth]{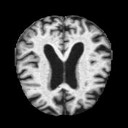}}
    \hfill
  \subfloat[Moderate Demented.\label{mod}]{%
        \includegraphics[width=0.23\linewidth]{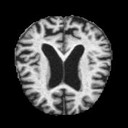}}
    \hfill 
  \subfloat[Non Demented.\label{nd}]{%
       \includegraphics[width=0.23\linewidth]{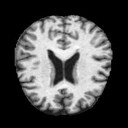}}
    \hfill
   \subfloat[Very Mild Demented.\label{vmd}]{%
       \includegraphics[width=0.23\linewidth]{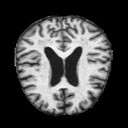}}
    \hfill

    \caption{Sample Images of Four Classes in the Dataset.}
    \label{sampleImages}
\end{figure}






\subsection{Image Pre-processing}
Pre-processing of data is one crucial process that prepares the dataset for corresponding classification model training. The resolution of MRI images of the dataset is 128 x 128, with a significant number of characteristics. To process high-resolution images high configured machine is necessary, which is expensive too. One of our targets is to observe the performance of low-resolution images. As a result, to reduce resource utilization for classifying these images, the resolution must be decreased. So, the MRI images are compressed to 32 x 32 in order to fit in the proposed approach.

\subsection{Feature Extraction}
Feature extraction is a leap in the dimension retrenchment procedure that splits and alleviates an initial amount of fresh data to more accessible classes. As a consequence, processing procedure will be easier. The minimization of data assists in the construction of the model with little machine activity, as well as the efficiency of the training and classification phases in the machine learning model. Due to this reason, researchers and the deep learning communities proposed several methods and techniques to decrease the data dimensionality for performance optimization. The Watershed Algorithm was used in this work as a feature-extraction approach to retrieve the features from the dataset.

\subsubsection{Watershed Algorithm}
Watershed algorithm is a well received segmentation technique \cite{kornilov2022review}, which provides numerous benefits for picture segmentation, such as ensuring closed section boundaries and producing reliable findings. The morphological watershed transform is a well-known image segmentation approach that uses mathematical morphology to divide a picture due to irregularities. Watersheds are defined by considering a picture in three dimensions: 2-dimensional co-ordinates against the intensity\cite{shaik2010feature}. The watershed algorithm developed on gathering certain foreground and background information, and then utilizing markers to run the watershed and discover the exact borders. This technique is useful for detecting contacting and overlapped objects in images. 

The marker based watershed transformation technique is an enhancement of the watershed method, which is employed to establish a marker in an picture. The marking can be a line, a spot, or a region; the placement is more significant than the indicated shape. Every tag presents a picture of a finalized separation, and choosing a mark is an important aspect in deciding segmentation. By adjusting the threshold, one can obtain all of the low values. With all these lowlands as a marker, the watershed transformation generates the final segmentation image\cite{zhang2011marker}. With the relation between its aims of breaking each small area, and in the middle of every target, the position of neighboring pixels with the same gray level of the formulation adds up to the local highest platform region strength, and noise is removed by filtration and image renovation. The seeding section is the largest target location. This seeding section is the extracted marker region. Image in grayscale of the destination within the marker contain a significant number of non seed areas with the noise or particular areas. The gray threshold is strongly tied to the tag area amount. The image with the highest gray level is from the decision's height. By deleting the marked portion of the regional encounters de-noising impact, the gray value is reduced. A targeted area of the markers is merged by obtaining the maximum gray level. The target particle size was chosen as a threshold image's gray level. If the gray threshold is too significant, all markers under the threshold mark of tiny target regions have been deleted. If the gray threshold is too minimum, extremely tiny particles have also been determined as a target. The watershed segmented picture is created after identifying the reproduced binary image.

\begin{figure}[htb]
    \centering
    \includegraphics[width=8.5cm,height=6.5cm]{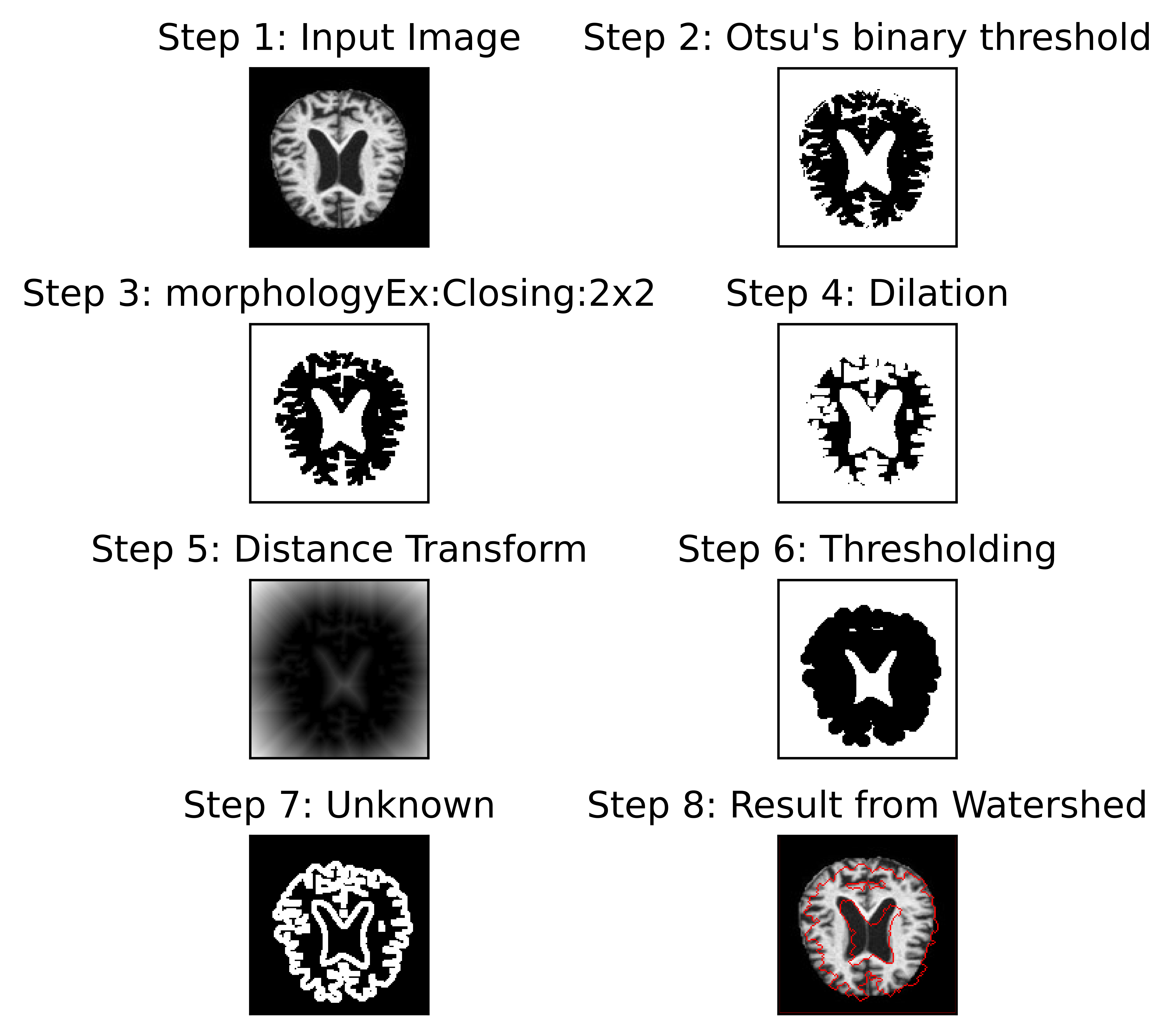}

    \caption{Steps of marker based watershed algorithm.}
    \label{watershedsteps}
\end{figure}

Figure \ref{watershedsteps} is showing the steps of the marker based watershed segmentation technique employed in this experimental work. These steps are expressed in the previous paragraph. For this research work, the generated image showd in step 7 of figure \ref{watershedsteps} is used to generate the features of the MRI images.

\subsection{Classification Models}
This section of the paper describes the machine learning models employed to identify the level of dementia in patients. For training, from ensemble algorithm, Random Forest (RF), from support vector networks, Support Vector Machine Classifier (SVM) and from neural network, convolutional neural network (CNN) approach is employed in this research work. The subsections provide a summary of the RF, SVM, and CNN models used to meet the clinical goal of dementia classification.

\subsubsection{Random Forest (RF)}
A group or ensemble of classifier and regression Trees is known as a Random Forest. In comparison to other machine learning methods, RF is regarded to be more steady in the existence of abnormalities and higher dimensional parameters spaces because it adheres to specified directives for tree building, self-testing, tree combination, and post-processing. It is also vigorous to overfitting \cite{sarica2017random}. In this research, we set the tree's maximum depth as 16 with zero random states.

\subsubsection{Support Vector Machine Classifier (SVM)}
One of the highest popular kernel based learning methods is the SVM technique, which was first developed by Vapnik and his team in the late 1970s and is notably popular in image classification tasks. SVM is a member of the class of supervised nonparametric algorithms, which are unaffected by the dispersion of the underlying data. That is one of the benefits of using SVMs over other statistical methods. SVM is a linear binary classifier that distinguishes just one border between two classes in its most basic configuration \cite{sheykhmousa2020support}. SVMs utilize a subset of the training specimen that is located nearest to the ideal decision boundary in the feature space, acting as support vectors, to increase the dissociation or border. In this research a polynomial kernel function is used with a degree value of 3.

\subsubsection{Convolution Neural Networks (CNN)}
In terms of image processing, CNN happens to be one of the finest and most widely used Deep Learning techniques. In this experimental work, the target is to use minimum layers and only three 2D Convolution layers are used. Fig \ref{cnn} demonstrates the CNN classification structure in this research.

\begin{figure}[htb]
    \centering
    \includegraphics[width=8.5cm,height=4.5cm]{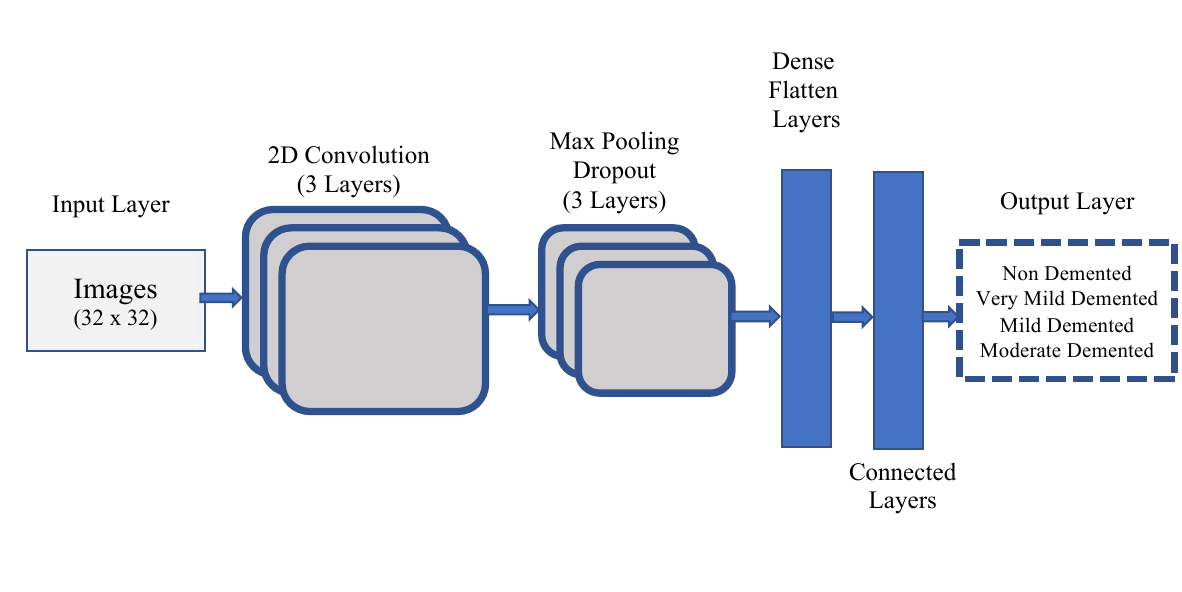}
    \caption{Architecture of CNN used in this research work.}
    \label{cnn}
\end{figure}

In the proposed CNN model, the count of convolutional layers and completely linked layers are three. The edges and color of the image are mostly captured by the first convolutional layer. As 2D images are used all the convolutional layers are two-dimensional. Explanation of the inputs, outputs, and amount of parameters that need to be adjusted for every layers of the model along three convolution layers are given below:

\par \textbf{First Layer}: Every gray scale image in the database is in 32x32 dimension. The first convolution layer receives this image as input, which is then processed using a single convolution filter. There are 1664 parameters in this layer that must be trained. Since each filter generates an image with a size of 28x28, the outcome of this layer is 28 x 28 x 1. 

\par \textbf{Second Layer}: The second convolutional layer utilizes an input with 64 filters and a size of 28x28x1. There are 73856 parameters throughout this layer that must be taught. As every filter produces an image of size 28x28 and every one of the 64 filters produces 64 pictures of size 28x28, the outcome of this layer is 28 x 28 x 1.

\par \textbf{Third Layer}: Images with 128 filters are used in the third layer. There are 65664 parameters in this layer that must be taught. As every filter produces an image with size 6x6 and the 128th filter produces 128 pictures of size 6x6, the product of this layer is 6 x 6 x 128.

\par \textbf{Output Layer}: The last layer utilizes the soft-max activation function, which has four possible outcomes, and is linked. This layer approximates the class values and assigns the four classes after receiving input from a fully linked layer. In this layer, 516 parameters have been modified. This classification indicates the class the subject belongs within.









\section{Experimental Result}\label{Res}
This segment of the article discusses the outcomes of the experiment and explains how they were conducted. We trained random forest, support vector machine classifier, and convolutional neural networks to perform four class classifications, such as Moderate Demented (MoD), Mild Demented (MD), Very Mild Demented (VMD), and Non-Demented (ND). The dataset comprises 6400 MRI images in total and is imbalanced. This study examines the efficacy of RF, SVM, and CNN with watershed and without watershed features for classifying dementia levels from MRI images. The dataset was split into 80\% and 20\% ratios for training and testing for RF and SVM evaluation. On the other hand, it is split into a 70\% training, 10\% validation, and test 20\% split ratio for the CNN evaluation. 

\subsection{Result Based on RF}
Throughout this research, the first step in classifying the four classes with the RF model and to determine the results using training, and test data. We have done it with and without the watershed feature of the image data. This model without watershed achieved an accuracy result of 91.25\% and with watershed 89.53\%. The confusion matrix of the random forest is stated in the figure \ref{RFcm}.

\begin{figure}[htb]
    \centering
  \subfloat[Without watershed.\label{a}]{%
       \includegraphics[width=0.47\linewidth]{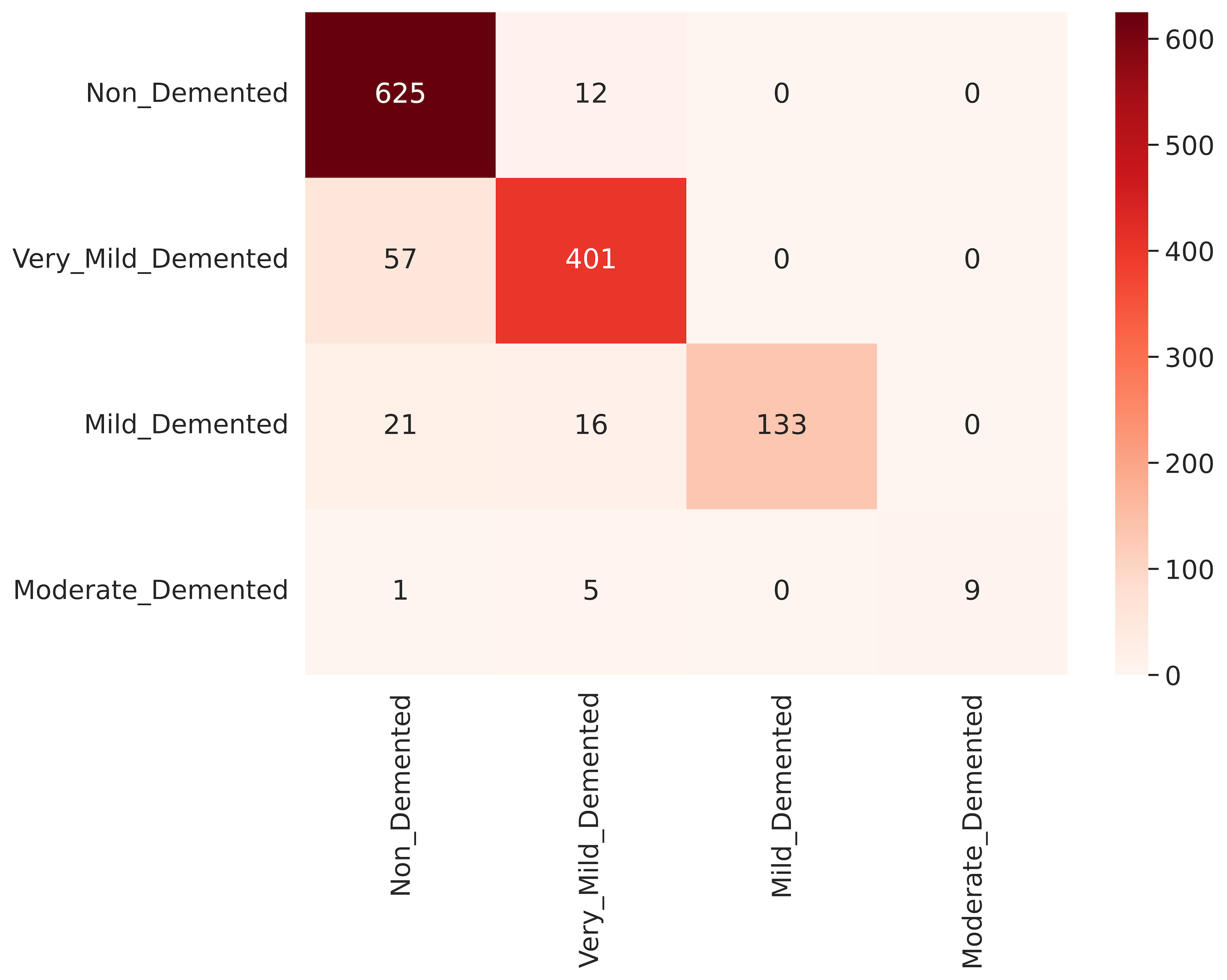}}
    \hfill
  \subfloat[With watershed\label{b}]{%
        \includegraphics[width=0.47\linewidth]{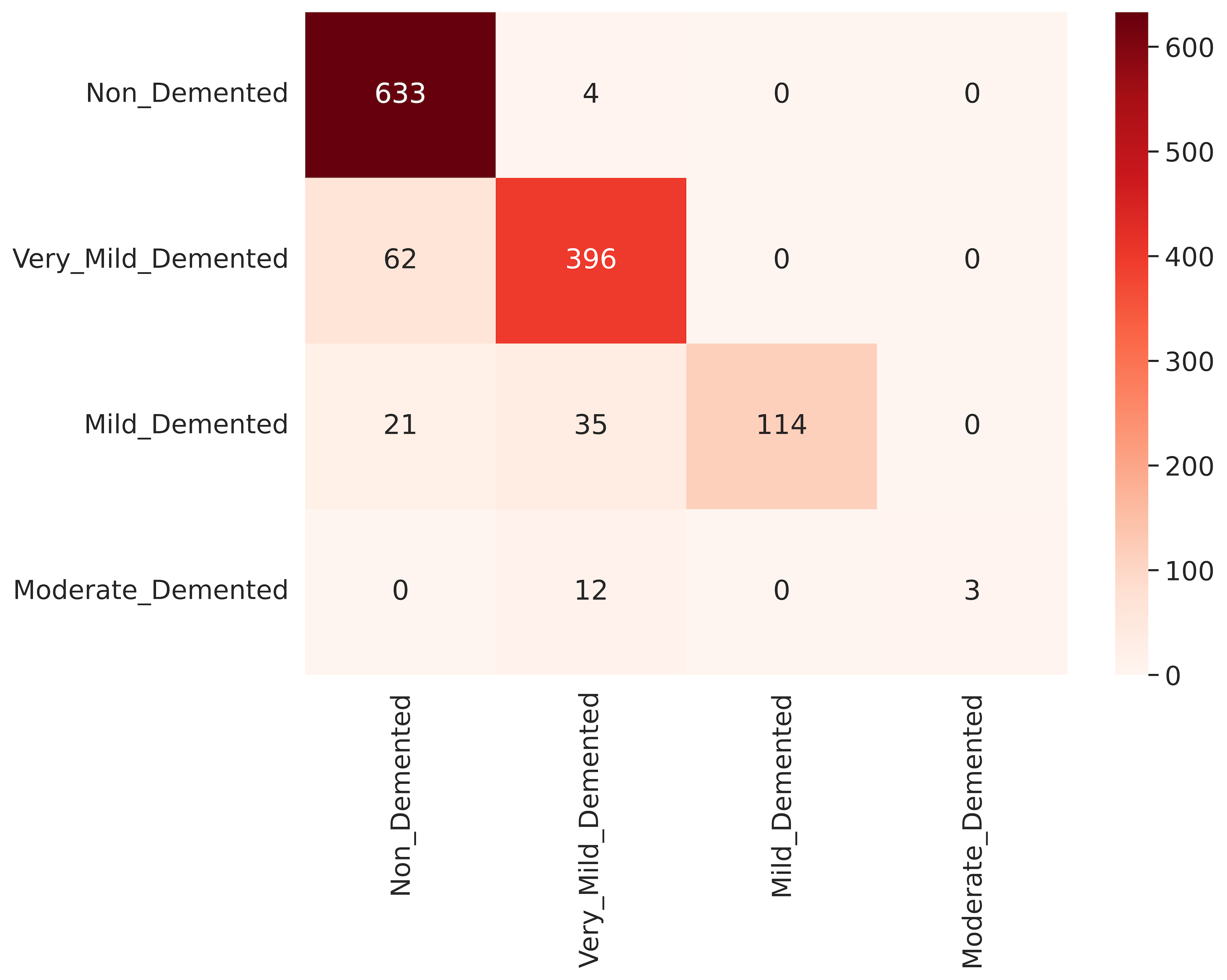}}
    \hfill 
    
  \caption{Confusion Matrix for Random Forest.}
  \label{RFcm} 
\end{figure}

Figure \ref{a} is showing the confusion matrix for random forests without watershed and figure \ref{b} is for those with the watershed. From these matrices, we can observe that though the overall accuracy is higher without watershed than with watershed, for the non-demented class the number of true positives is greater with the watershed. For other classes, the number of true positives is higher without watershed. For the non-demented class number of images was maximum. We can say that, if the dataset were balanced and the other three classes had more images than with watershed the RF could have given higher accuracy.

\subsection{Result Based on SVM}
Support vector machine classifier without watershed achieved an accuracy result of 80.70\% and with watershed 96.25\%. The confusion matrix of support vector machine classifier is stated in figure \ref{SVMcm}.

\begin{figure}[htb]
    \centering
  \subfloat[Without watershed.\label{1a}]{%
       \includegraphics[width=0.47\linewidth]{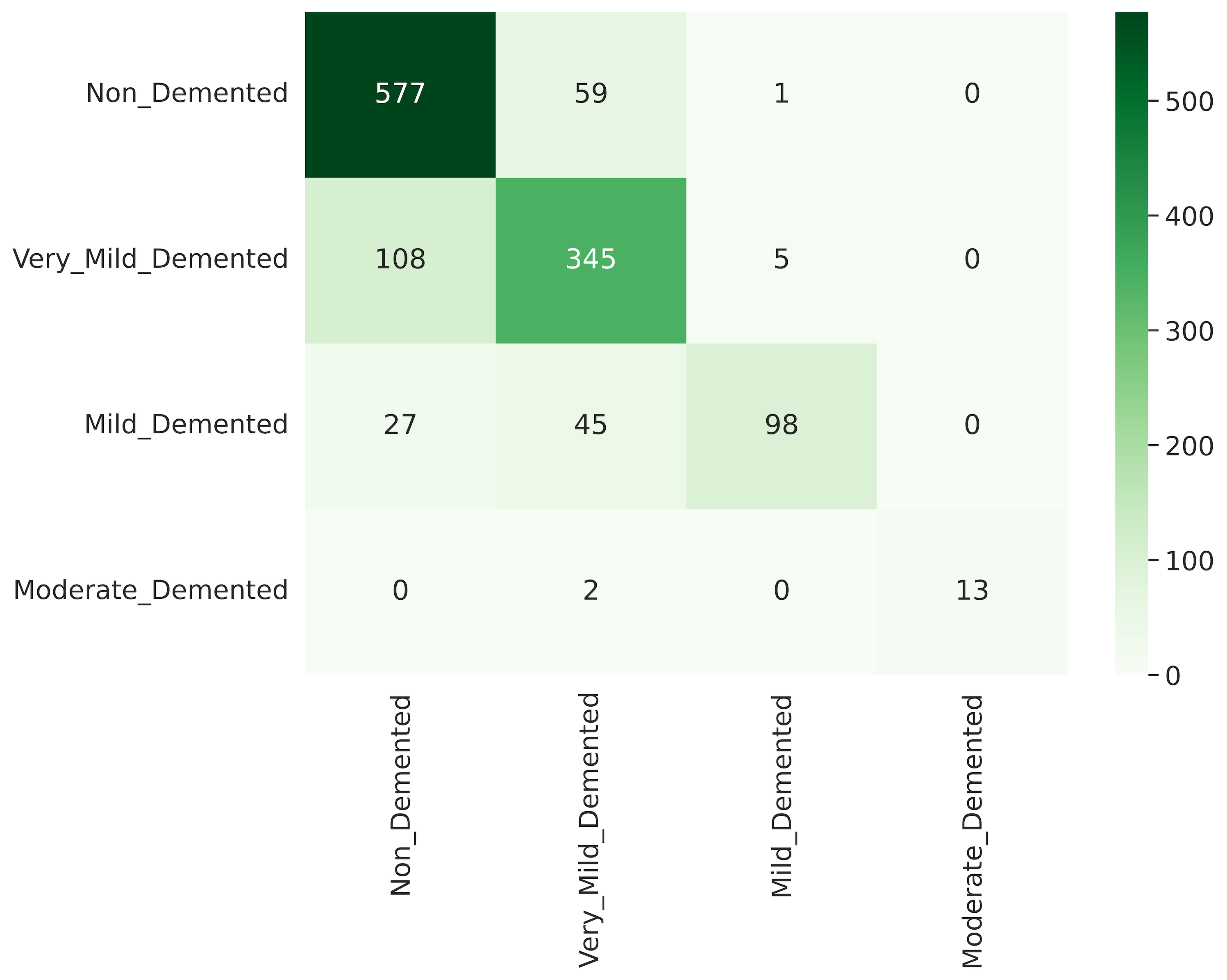}}
    \hfill
  \subfloat[With watershed\label{1b}]{%
        \includegraphics[width=0.47\linewidth]{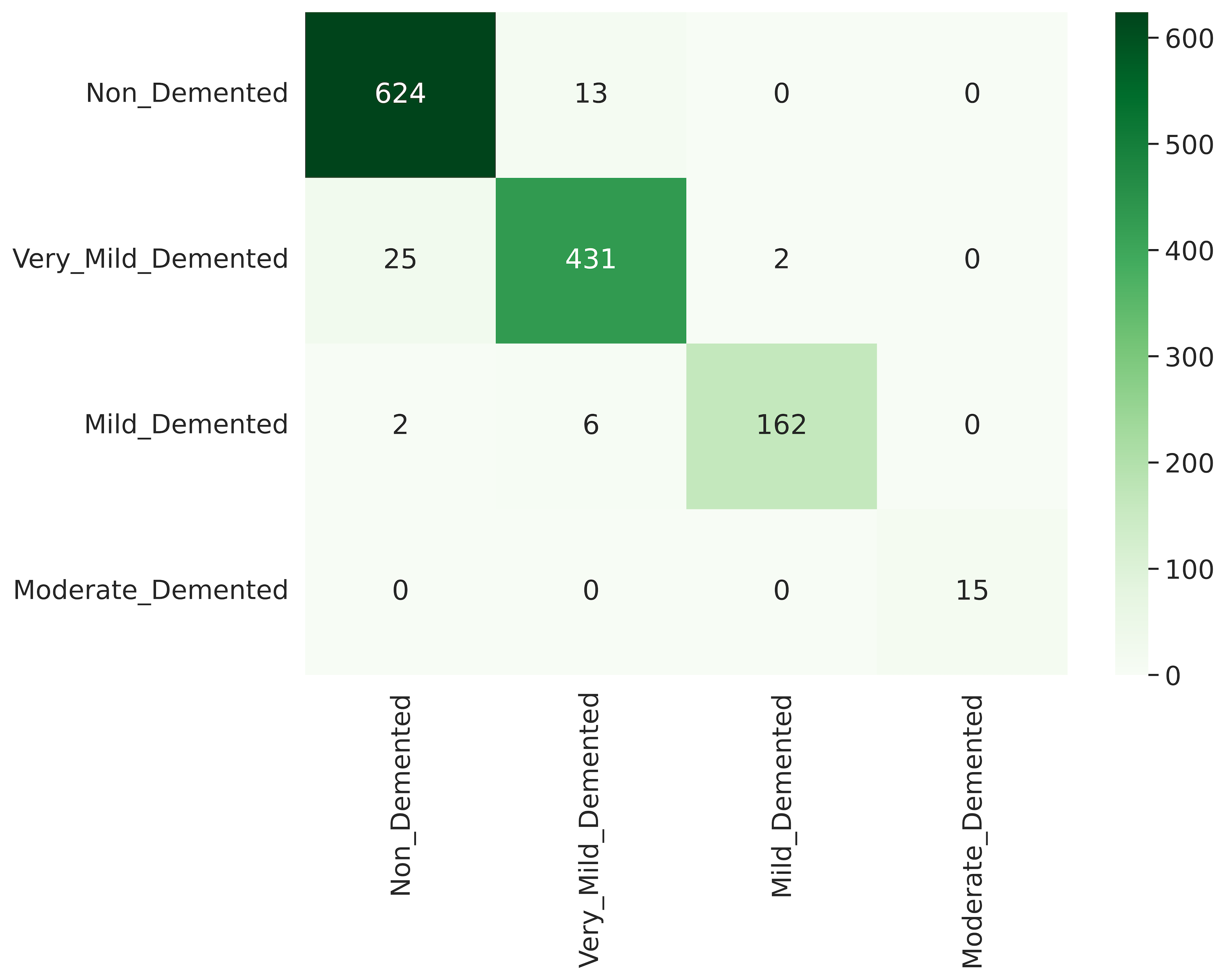}}
    \hfill 
    
  \caption{Confusion Matrix for Support Vector Machine Classifier.}
  \label{SVMcm} 
\end{figure}

Figure \ref{1a} is showing the confusion matrix of support vector machine without watershed and figure \ref{1b} is for those with the watershed. From these, we can observe that for all the classes the number of true positives is higher with the watershed than without the watershed. The difference is notable and for this reason, the overall accuracy with watershed is much higher than without watershed. We can say that, if the dataset were balanced and the other three classes had more images than with watershed the SVM could have given a much better performance.

\subsection{Result Based on CNN}
The CNN framework was trained and validated using 50 epochs to classify the four levels of dementia. The graph of training \& validation accuracy without watershed is given in the figure \ref{TVaccWoWS} and figure \ref{TVaccWWS}. The training \& validation loss graph with watershed is demonstrated in figure \ref{TVlossWoWS} and figure \ref{TVlossWWS}.  In figure \ref{CNNtvAccLoss}, it can be noticed that without watershed the training accuracy is around 92\% and loss is around 10\% and validation accuracy is around 94\% and loss is around 7\%. But with watershed, the training accuracy is around 98\% and loss around 3\% and validation accuracy are around 94\% and loss is around 8\%. From figure \ref{CNNtvAccLoss}, for all the graphs we can observe that the opening among training accuracy and validation accuracy or between training loss and validation loss is not so large. So, 50 epochs are enough for training and validating our models with and without the watershed. From figure \ref{TVaccWoWS}, we can see that after 40 epochs the accuracy for training and validation is in a stable position without watershed but in figure \ref{TVaccWWS}, both the training and validation accuracy are stable after around 25 epochs. So we can say the watershed algorithm helped the model to be trained and validate with fewer epochs.

\begin{figure}[htb]
    \centering
  \subfloat[Training and Validation Accuracy Without watershed.\label{TVaccWoWS}]{%
       \includegraphics[width=0.47\linewidth]{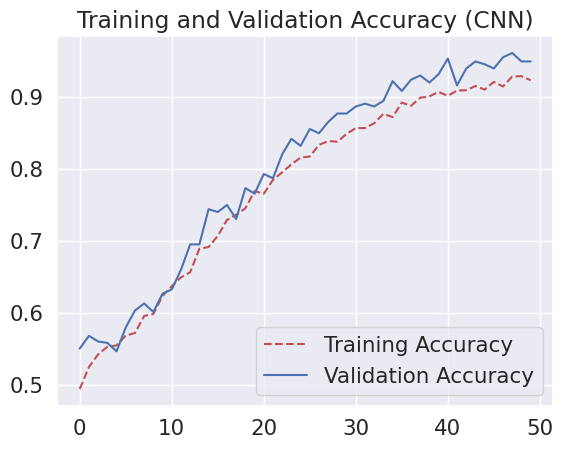}}
    \hfill
  \subfloat[Training and Validation Accuracy With watershed\label{TVaccWWS}]{%
        \includegraphics[width=0.47\linewidth]{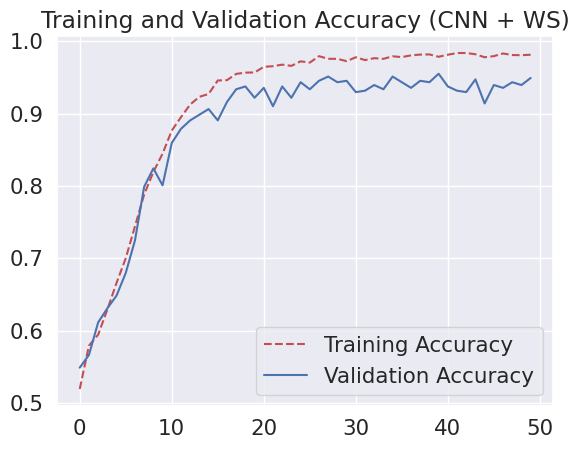}}
    \hfill 
    \subfloat[Training and Validation Loss Without watershed.\label{TVlossWoWS}]{%
       \includegraphics[width=0.47\linewidth]{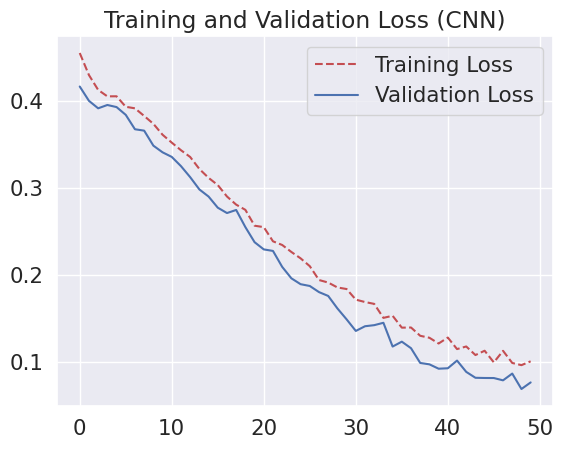}}
    \hfill
  \subfloat[Training and Validation Loss With watershed\label{TVlossWWS}]{%
        \includegraphics[width=0.47\linewidth]{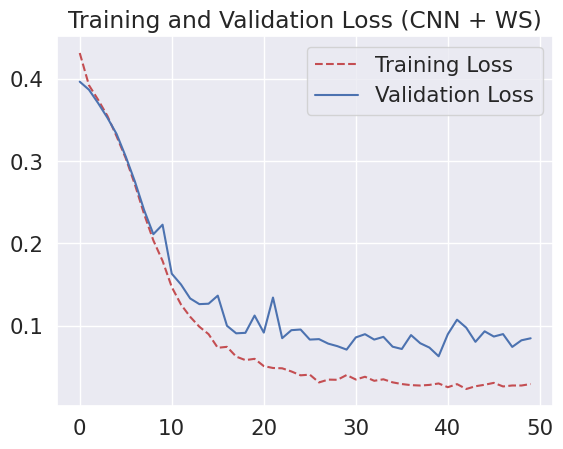}}
    
  \caption{Training \& Validation Accuracy and Loss of the CNN.}
  \label{CNNtvAccLoss} 
\end{figure}

After the system has been trained, it is evaluated using the testing set in which the images that are present were not given to the system throughout training. The CNN model achieves accuracy results of 93.98\% without watershed feature, and 95.16\% with watershed respectively for the test set. The projected class and labeled classes from the four separate categories are plotted on the confusion matrix. The model effectiveness over the training dataset is provided by the confusion matrix. Figure \ref{CNNcm} depicts the confusion matrix used by this structure to categorize the phases of dementia. 

\begin{figure}[htb]
    \centering
  \subfloat[Without watershed.\label{cmCNN1}]{%
       \includegraphics[width=0.47\linewidth]{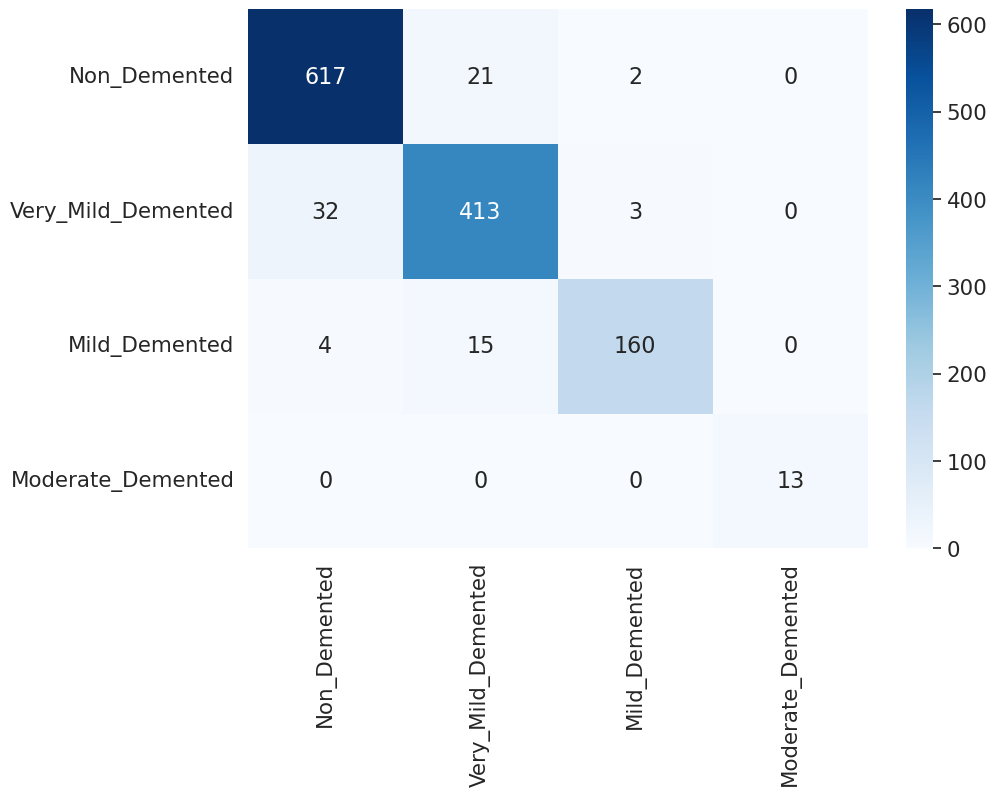}}
    \hfill
  \subfloat[With watershed\label{cmCNN2}]{%
        \includegraphics[width=0.47\linewidth]{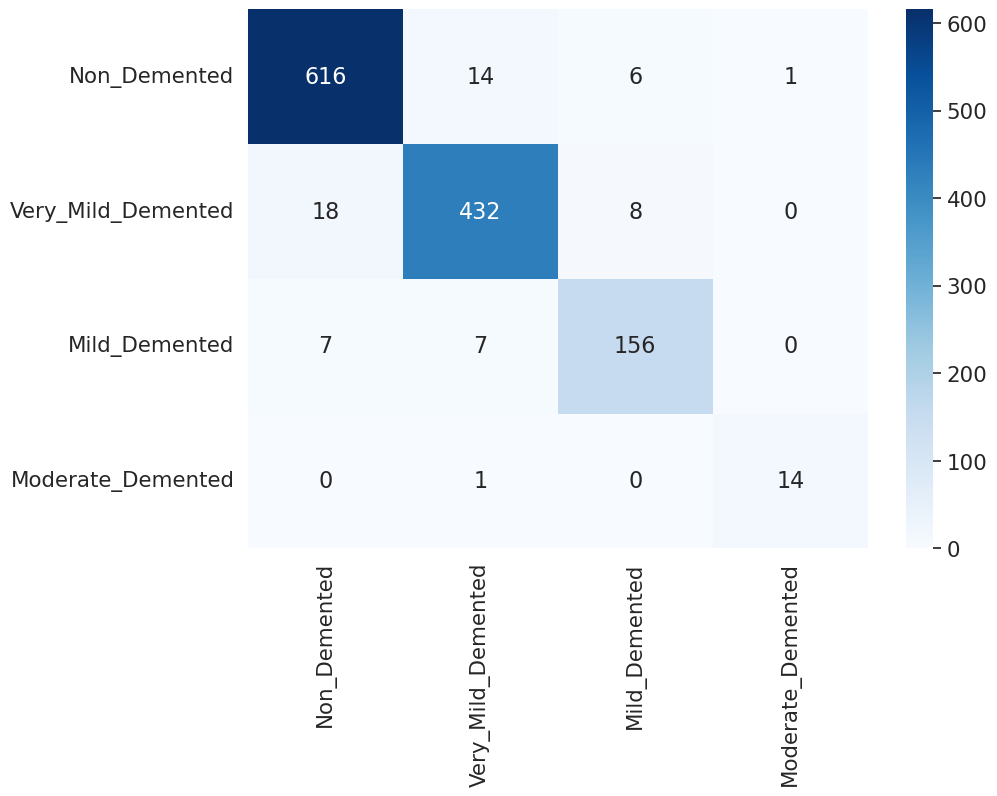}}
    \hfill 
    
  \caption{Confusion Matrix for Convolutional Neural Network.}
  \label{CNNcm} 
\end{figure}

Figure \ref{cmCNN1} is showing the confusion matrix for CNN without watershed and figure \ref{cmCNN2} is for those with the watershed. From these matrices, we can observe that though the overall accuracy is higher with watershed than without watershed, for the non demented \& mild demented class the number of true positives is fewer with the watershed but the difference is very low. For other classes, the number of true positives is higher with watershed and the difference is high. So we can say that if the dataset were balanced and larger, then with watershed the CNN could have given much higher accuracy.





\subsection{Comparisons between RF, SVM and CNN}
In this research work, the performances metrics, such as recall, precision, and f1-score were also measured for all the models explored. The all-inclusive classification performance of the proposed models is demonstrated in table \ref{table:Accuresult}. It compares the performance of respective RF, SVM, and CNN for the four classes classification on the dataset used in this research work. 

\begin{table}[htb!]
\centering
\caption{Accuracy, Precision, Recall and F1-Score Result of Models Used}

\def\arraystretch{1.5} 

\begin{tabular}{| p{1.4cm} | p{1.3cm} | p{1.3cm} | p{1.2cm}  | p{1.4cm}  |} 
 \hline
Model & Accuracy \%   & Precision  \% & Recall  \% & F1-Score  \%  \\ 

\hline
RF & 91.25 & 92.0 & 91.0 & 91.0 \\
\hline
WS $+$ RF & 89.53 & 90.0 & 90.0 & 89.0 \\
\hline
SVM & 80.70 & 81.0 & 81.0 & 80.0 \\
\hline
\textbf{WS $+$ SVM} & \textbf{96.25} & \textbf{98.0} & \textbf{97.0} & \textbf{97.0}  \\
\hline
CNN & 93.98 & 94.0 & 94.0 & 94.0 \\
\hline
WS $+$ CNN & 95.16 & 95.0 & 95.0 & 95.0 \\
\hline

\end{tabular}

\label{table:Accuresult}
\end{table}

From the table \ref{table:Accuresult}, it can be observed that, with the high classification effectiveness in terms of accuracy, precision, recall, and f1-score, the SVM model with watershed feature outperforms all the other models used in this research work. We can observe a little high precision for every model compared to the accuracy which is useful for these different class classifications from the table \ref{table:Accuresult}. Higher precision means the possibility of predicting an image to the correct class is high. It will be more helpful for balanced and larger datasets. We can also observe that though the resolution of the MRI images is 32 x 32 which is very low, the accuracy for all the models is greater than 80 percent. It is also acceptable that with low-resolution MRI images, which will consume fewer resources, it is possible to identify dementia levels.

On the other hand, among the three models for SVM and CNN, using the watershed algorithm as a feature increased the accuracy and performance of the models. For SVM the difference is very high and with watershed, it is giving the best performance. But RF using watershed the performance decreased slightly. This might be due to the imbalanced and small dataset. For CNN, generally, a larger dataset is very useful but for training the SVM small dataset is also useful \cite{9132851}. In our research, the dataset contains only 6400 images, which is very small to train CNN properly. But we are getting good performance for this dataset too, which is around 94 percent without watershed and 95 with the watershed. This performance with the watershed can be increased by extending and balancing the dataset size. So, we can tell that the watershed segmentation algorithm might contribute as a feature in Alzheimer's disease classification using MRI images.

\subsection{Performance Comparison with Some Other Works}
The mentioned method is affirmed by contrasting the results to earlier research that focused on the earlier diagnosis of Alzheimer's disease using a classification system of four classes, as shown in the table \ref{table:ResCompariWOtherWorks}. The suggested SVM model performs superior in terms of accuracy, precision, and F1-score, classifying the four types of dementia with 96.25\% accuracy, 98.0\% precision, and 97\% F1 score.

\begin{table}[htb!]
\centering
\caption{Results Comparison with Existing Works}

\def\arraystretch{1.5} 

\begin{tabular}{| p{1cm} | p{1.7cm} | p{1cm} | p{0.9cm} | p{0.7cm} | p{0.8cm} |} 

 \hline
Reference & Methodology & Accuracy &  Precision & Recall & F1-Score\\ 

\hline
This Study & \textbf{WS $+$ SVM} & \textbf{96.25} & \textbf{98.0} & 97.0 & \textbf{97.0}\\
\hline
Article \cite{de2023prediction} & CNN & 89.0 & 89.0 & 89.0 & 89.0 \\
\hline
Article \cite{bharanidharan2022modified} & HT-MGWRO & 93.16 & 90.74 & 94.23 & 92.45 \\
\hline
Article \cite{bharati2022dementia} & Voting 1 (RF, GB, XGB) & 92.0 & 91.0 & 91.0 & 91.0 \\
\hline
Article \cite{herzog2021brain} & SVM & 92.5 & 92.0 & 93.0 & 92.49 \\
\hline
Article \cite{mohammed2021multi} & Random Forest & 94.0 & 93.0 & 98.0 & 95.43 \\
\hline
Article \cite{prajapati2021efficient} & DNN & 85.19 & 76.93 & 72.73 & 74.77 \\
\hline
Article \cite{bansal2020classification} & BOF + SVM & 93.0 & NA & NA & NA \\
\hline
Article \cite{bidani2019dementia} & DCNN & 81.94 & NA & NA & NA \\
\hline
\end{tabular}

\label{table:ResCompariWOtherWorks}
\end{table}

\section{Conclusion}\label{con}

In this study, a methodology is proposed for classifying Alzheimer's disease (AD) dementia levels using feature extraction with the watershed technique in combination with RF, SVM, and CNN architectures. The standard ADNI data is utilized to train and assess the models for classifying the four stages of dementia. Results indicate that SVM with watershed features outperforms other models on the ADNI dataset, achieving an impressive overall accuracy of 96.25\% when tested with data from all four dementia levels. This approach effectively identifies brain areas connected with AD and serves as a valuable decision supporting system for doctors to estimate the seriousness of AD grounded on dementia level. However, it should be noted that the dataset used in this research has a class imbalance issue, which was not addressed in this study. Future work could focus on addressing this limitation to further enhance the model's performance. This proposed approach has the potential to be employed as a standalone framework for dementia level evaluation and AD diagnosis support in clinical settings, and can be further validated on diverse datasets in future studies.

 

\ifCLASSOPTIONcaptionsoff
  \newpage
\fi

\bibliographystyle{IEEEtran}

\bibliography{ref.bib}

\end{document}